\documentclass[letterpaper, 10 pt, conference]{ieeeconf}  

\IEEEoverridecommandlockouts                              

\usepackage{algorithmic}
\usepackage{rotating}
\usepackage{graphicx}
\usepackage{textcomp}
\usepackage[ruled, algo2e]{algorithm2e}
\usepackage{xcolor}
\usepackage{amsmath,amssymb,amsfonts}

\title{\LARGE \bf
Compact NSGA-II for Multi-objective Feature Selection*
}

\author{Sevil Zanjani Miyandoab$^{**1}$, Shahryar Rahnamayan$^{**2}$, SMIEEE, Azam Asilian Bidgoli$^{**3}$
\thanks{*This research is supported by the Vector Scholarship in Artificial
Intelligence, provided through the Vector Institute.}
\thanks{**Nature-Inspired Computational Intelligence (NICI) Lab}
\thanks{$^1$Department of Electrical, Computer, and Software Engineering, Ontario Tech University, Oshawa, 
ON, Canada
        {\tt\small sevil.zanjanimiyandoab@ontariotechu.net}}%
\thanks{$^2$Department of Engineering, Brock University, St. Catharines, ON, Canada
        {\tt\small srahnamayan@brocku.ca}}%
\thanks{$^3$Faculty of Science, Wilfrid Laurier University, Waterloo, ON, Canada
        {\tt\small Abidgoli@wlu.ca}}%
}

\graphicspath{ {./images/} }

\usepackage{tikz}
\usepackage{textcomp}
\usepackage{hyperref}
\usepackage{lipsum}

\newcommand\copyrighttext{%
  \footnotesize \textcopyright 2023 IEEE. Personal use of this material is permitted.
  Permission from IEEE must be obtained for all other uses, in any current or future
  media, including reprinting/republishing this material for advertising or promotional
  purposes, creating new collective works, for resale or redistribution to servers or
  lists, or reuse of any copyrighted component of this work in other works.
  DOI: \href{https://ieeexplore.ieee.org/abstract/document/10394458}{10.1109/SMC53992.2023.10394458}}
\newcommand\copyrightnotice{%
\begin{tikzpicture}[remember picture,overlay]
\node[anchor=south,yshift=10pt] at (current page.south) {\fbox{\parbox{\dimexpr\textwidth-\fboxsep-\fboxrule\relax}{\copyrighttext}}};
\end{tikzpicture}%
}

\begin{document}

\maketitle
\copyrightnotice
\thispagestyle{empty}
\pagestyle{empty}

\begin{abstract}

Feature selection is an expensive challenging task in machine learning and data mining aimed at removing irrelevant and redundant features. This contributes to an improvement in classification accuracy, as well as the budget and memory requirements for classification, or any other post-processing task conducted after feature selection. In this regard, we define feature selection as a multi-objective binary optimization task with the objectives of maximizing classification accuracy and minimizing the number of selected features. In order to select optimal features, we have proposed a binary Compact NSGA-II (CNSGA-II) algorithm. Compactness represents the population as a probability distribution to enhance evolutionary algorithms not only to be more memory-efficient but also to reduce the number of fitness evaluations. Instead of holding two populations during the optimization process, our proposed method uses several Probability Vectors (PVs) to generate new individuals. Each PV efficiently explores a region of the search space to find non-dominated solutions instead of generating candidate solutions from a small population as is the common approach in most evolutionary algorithms. To the best of our knowledge, this is the first compact multi-objective algorithm proposed for feature selection. The reported results for expensive optimization cases with a limited budget on five datasets show that the CNSGA-II performs more efficiently than the well-known NSGA-II method in terms of the hypervolume (HV) performance metric requiring less memory. The proposed method and experimental results are explained and analyzed in detail.

\end{abstract}

\section{INTRODUCTION}

Feature selection refers to eliminating as many features as possible without sacrificing classification accuracy, or the accuracy of any other post-processing method. Due to the recent surge in real-world datasets' dimensionality to thousands and millions of features, feature selection methods have become increasingly crucial \cite{chandrashekar2014survey}. Removing irrelevant and redundant features reduces the curse of dimensionality for methods in which these features are inputs to their models. Therefore, it enhances the performance of a classifier and reduces computational requirements, and also makes the model easier to visualize and/or interpret \cite{chandrashekar2014survey,ghalwash2016structured}. 

However, eliminating too many features leads to degrading classification accuracy. Therefore, these two conflicting objectives in multi-objective feature selection, the number of features and classification accuracy, should be optimized simultaneously \cite{bidgoli2020evolutionary}. Multi-objective metaheuristic algorithms, including evolutionary algorithms, can effectively solve these kinds of problems.


In recent years, many multi-objective algorithms have been proposed to enhance feature selection. Hamdani et al. \cite{hamdani2007multi} have investigated the performance of NSGA-II (Non-dominated Sorting Genetic Algorithm II) \cite{deb2002fast} on multi-objective feature selection of five datasets from the UCI repository. Jiao et al. \cite{jiao2022solving} have introduced a multi-objective optimization method for feature selection in classification, named PRDH. This algorithm includes a duplication handling method to enhance the diversity of the population in the objective and search spaces and a novel constraint-handling method to select feature subsets that are more informative and strongly relevant. Cheng et al. \cite{cheng2022variable} have proposed a variable granularity search-based multi-objective evolutionary algorithm, VGS-MOEA, which significantly reduces the search space of large-scale feature selection problems. In this method, each bit of the solutions represents a feature subset. This subset (granularity) is larger at the beginning and refined gradually to find higher-quality features.

In a feature selection task, each evaluation call consists of training a classifier on the subset of selected features. Accordingly, increasing dimensionality makes the problem highly costly, and algorithms that produce a satisfactory result in a given budget are valuable. The compact genetic algorithm (CGA), originally introduced by Harik et al.\cite{harik1999compact}, exemplifies such an algorithm. In this algorithm, instead of maintaining a set of candidate solutions as a population, the probability distribution of them is stored. CGA has the same operational behavior as a simple GA with a uniform crossover while using less memory. 

Generally, compactness makes population-based algorithms suitable for hardware contexts with limited computation power, such as micro-controllers and commercial robots. Several studies \cite{aporntewan2001hardware, gallagher2004family, jewajinda2008cellular, mininno2008real} have employed different CGA schemes on embedded hardware.

Mininno et al. \cite{mininno2008real} also presented a real-valued CGA. In this method, PV consists of two vectors for saving the mean and standard deviation of each real-valued feature. In this way, a Gaussian probability distribution function (PDF) for each variable can be obtained. In each iteration, a new individual is generated using these two vectors and compared with the elite, and the vectors are updated using a rule.
Iacca et al. \cite{iacca2019compact, iacca2020re} proposed a novel restart mechanism from a partially random individual, to overcome the drawbacks of compact algorithms, e.g., premature convergence and weak performance on non-separable problems. Their method, called Re-Sampled Inheritance, produces promising results when combined with compact algorithms.

To the best of our knowledge, there is only one multi-objective compact algorithm; Velazquez et al.\cite{velazquez2014multi} introduced the first compact multi-objective algorithm based on differential evolution inspired by \cite{mininno2010compact}. They use only one probability vector to represent the population and generate only one new solution in each iteration. We think it may limit Pareto front solutions' distribution. Although their results are competitive with several multi-objective evolutionary algorithms, the distribution of the solutions is not better than NSGA-II. This method has been tested on ZDT and DTLZ benchmarks. 

In this study, we propose a compact version of the well-known multi-objective evolutionary algorithm, NSGA-II. 
This novel method, called CNSGA-II, is then employed for multi-objective feature selection, and its performance is compared with classical binary-encoded NSGA-II.



Using our proposed method, $N$ distinct PVs are stored to cover different areas of the search space and enhance solution distribution. Unlike previous related works, this approach can efficiently identify and explore potential areas of the search space. This happens without consuming too much memory to save a large population of individuals. A non-dominated sorting algorithm preserves the best set of features or individuals. During the optimization process, PVs become increasingly focused on $N$ top members of the upper front(s) with more crowding distances. This algorithm outperforms NSGA-II in terms of the hypervolume (HV) performance measure, which is a well-known metric to evaluate convergence rate and solution distribution on the optimal Pareto front.

This paper is structured as follows. Section II provides a comprehensive background review, offering an overview of the relevant literature and theoretical foundations. Building upon this background knowledge, Section III introduces our proposed compact NSGA-II algorithm, highlighting its key features and improvements over existing approaches. In Section IV, we present the experimental results and analysis obtained from applying the proposed algorithm to a set of datasets and discuss its performance. Finally, in Section V, we summarize our findings and suggest potential directions for future research.

\section{BACKGROUND REVIEW}

The main components of the proposed method are CGA and multi-objective optimization which are explained in detail in the ensuing subsections.  

\subsection{Compact Population-based Algorithms }
 
Harik et al. \cite{harik1999compact} introduced the compact strategy for the first time in the compact genetic algorithm (CGA). After initializing the PV as a vector of 0.5s with the length of $|features|$, their method picks two random individuals from the probability vector (PV) in each iteration. To this end, it generates a vector consisting of random numbers in the interval $[0,1]$ with a uniform distribution for each random individual. It compares each element of the randomly generated vector with the corresponding element in PV. If $random[i] < PV[i]$, then $new\: individual[i] = 1$, otherwise $new\: individual[i] = 0$. After comparing the objective value of the new individuals, the PV elements are updated according to their non-similar genes. If the bit number $i$ of the winner has the value of 1 and the same bit of the loser has the value of 0, the corresponding element of PV would be increased by $1/n$ (n is the population size), and vice versa. 
This can be considered equivalent to steady-state binary tournament selection, in which the proportion of the winning gene increases by $1/n$. Similarly, to update each PV element, we increase a gene’s value’s proportion by a small step of $1/n$.

In the simple binary genetic algorithm (GA), we need to store $2 \times n$ (two times population size) bits for each gene position; however, the CGA keeps the proportion of ones (and zeros) using $log_2 (n+1)$ bits for each variable as an element of the Probability Vector (PV), where $n$ indicates the virtual population size or $1/StepSize$. Therefore, this algorithm could serve as an advantageous choice for feature selection, especially  when dealing with large- or huge-scale datasets. 

Ahn et al. \cite{ahn2003elitism} proposed several modifications and utilized elitism in the CGA in two redesigned schemes of the CGA: persistent elitist compact genetic algorithm (pe-CGA), and non-persistent elitist compact genetic algorithm (ne-CGA). One of the main problems of the CGA is the possible loss of the best current solution due to lack of memory. These two elitism-based CGA algorithms generate one individual in each iteration while keeping the elite, and update PV based on their non-similar genes. pe-CGA keeps the current best solution until a better solution is found. By reducing the pe-CGA elitism pressure, the ne-CGA reduces the likelihood of premature convergence. 
 Their results show that these two algorithms generate better solution sets and have higher convergence speeds than GA and CGA without compromising memory or computational requirements. 

An extended CGA (ECGA) is presented in \cite{harik1999linkage, harik2006linkage} based on the relation between the linkage-learning problem and learning probability distributions, and \cite{sastry2000extended} analyzes its behavior comprehensively. Compact schemes for many other search algorithms have also been developed. Mininno et al. \cite{mininno2010compact} proposed compact differential evolution (CDE) methods by employing statistical description rather than an entire population of solutions. They show that CDE performs efficiently and fast with limited memory requirements. 

Neri et al. \cite{neri2013compact} presented compact particle swarm optimization (CPSO), which uses a probabilistic representation of the swarm’s behavior, instead of storing positions and velocities. Zheng et al. \cite{zheng2021compact} applied the compact strategy to the adaptive particle swarm algorithm and proposed a compact adaptive particle swarm method (CAPSO). Liu et al. \cite{liu2022novel} introduced a novel compact particle swarm optimization algorithm, which uses a Pareto distribution to describe particle swarm positions.

Besides memory efficiency, the compact strategy is based on sampling; thus, it can be beneficial in certain situations, such as solving huge-scale problems, which offer linear complexity compared to common high-order complicated algorithms.

\subsection{Multi-objective Optimization}
Multi-objective optimization is defined as optimizing two or more conflicting objectives. The algorithms to solve these problems usually make a trade-off decision and generate a set of solutions instead of a single one. This set of solutions is called the Pareto front, or non-dominated solutions. The Pareto front is created using the concept of dominance, which is used to compare multi-objective candidate solutions. 

\textbf{Definition 1. Multi-objective Optimization}~\cite{asilian2022machine}
\begin{eqnarray}
\begin{aligned}
& Min/Max\  F(\pmb x)=[f_{1}(\pmb x),f_{2}(\pmb x),...,f_{M}(\pmb x)] \\ 
&s.t. \quad L_{i}\leq x_{i}\leq U_{i}, i=1,2,...,d
 \end{aligned}
\end{eqnarray}
Subject to the following equality and/or inequality constraints.
\begin{eqnarray}
\begin{aligned}
g_j(\pmb x) \leq 0 \quad j=1,2,...,J\\
h_k(\pmb x)=0 \quad k=1,2,...,K
\end{aligned}
\end{eqnarray}
where $M$ is the number of objectives, $d$ is the number of decision variables (i.e., dimension), and the value of each variable, $\pmb x_{i}$, is in the interval $[L_{i}, U_{i}]$. $f_{i}$ represents the objective function, which should be minimized or maximized.

One of the commonly used  concepts for comparing candidate solutions in such problems is dominance.

\textbf{Definition 2. Dominance Concept} \\
If  $\pmb x=(x_{1},x_{2},...,x_{d})$ and  $ \acute{\pmb x}=(\acute{x}_{1},\acute{x}_{2},...,\acute{ x}_{d})$ are two vectors in a minimization problem search space, $\pmb x$ dominates $\acute{\pmb x}$ ($\pmb x\prec\acute{\pmb x}$) if and only if
\begin{eqnarray}
\begin{aligned}
&\forall i\in{\{1,2,...,M\}}, f_i(\pmb x)\leq f_i(\acute{\pmb x}) \wedge\\ 
&\exists j \in{\{1,2,...,M\}}: f_j(\pmb x)<f_j(\acute{\pmb x})
\end{aligned}
\end{eqnarray}
This concept defines the optimality of a solution in a multi-objective space. Candidate solution $\pmb x$ is better than $\acute{\pmb x}$ if it is not worse than $\acute{\pmb x}$ in any of the objectives and at least it has a better value in one of the objectives. All of the solutions, which are not dominated by any other solution, create the Pareto front and are called non-dominated solutions \cite{bidgoli2021reference}.

Another metric that is commonly used for comparing candidate solutions in multi-objective problems is crowding distance. For computing the crowding distance of the individuals, we sort them based on each objective and assign an infinite distance to the boundary individuals (minimums and maximums). Other individuals are assigned the sum of the normalized Euclidean individual distance from their neighbors with the same rank on all objectives. This estimates the importance of an individual in relation to the density of individuals surrounding it~\cite{deb2002fast}.

 Multi-objective algorithms attempt to find the Pareto front by utilizing generating strategies/operators and selection schemes. The non-dominated sorting (NDS) algorithm~\cite{deb2002fast} is one of the popular selection strategies based on the dominance concept. It ranks the population's solutions in different levels of optimality. The algorithm starts with determining all non-dominated solutions at the first rank.
 
 In order to identify the second rank of individuals, the non-dominated vectors are removed from the set to process the remaining candidate solutions in the same way. This process will continue until all individuals are grouped into different Pareto levels. Deb et al. \cite{deb2002fast} has introduced one of the most famous and impressive state-of-the-art multi-objective optimization methods, NSGA-II, based on the mentioned dominance concept. It is the main competitor to our proposed method.

\section{PROPOSED COMPACT NSGA-II}

As mentioned earlier, feature selection is aimed at improving classification accuracy and reducing the number of features. To calculate classification accuracy and assess the subset of selected features, a classifier should be trained and evaluated on the training set during the optimization phase. After the optimization process, the final set of features (Pareto front) will be evaluated using the test set. 

In order to design a minimization problem, we define classification error as the first objective using the classifier's predictions:

\begin{equation}
\label{f1}
\mathit{Classification \:Error}= 1 - \frac{\# \:Correct\: Predictions}{Total \:\# Predictions}
\end{equation}

Minimizing the ratio of selected features is the second objective. To this end, we count the number of variables with the value of 1 for each candidate solution which represents the ratio of selected features.

\begin{equation}
\label{f2}
\mathit{Ratio\: of\: Selected\: Features}=\frac{Number \:of \: 1 s }{Total \:number\: of \:Features }
\end{equation}

In this way, both objectives will have real values in the interval \([0, 1]\).

PV is a probability vector to preserve solution distribution. In fact, each variable is a real value between 0 and 1 representing the probability of selection of each feature. The algorithm starts with generating $N$ number of PVs. Then, at each iteration, $N$ individuals are generated using these PVs (i.e., sampled from the distribution) and merged with the previous population. To find the Pareto front, we apply NDS to all existing individuals. As part of the PV update, $N$ top solutions are selected from the population as leaders based on NDS and crowding distance.  

CNSGA-II consists of several major steps as mentioned in Algorithm~\ref{alg-one}. The details of each step are provided as follows:

\begin{enumerate}
   \item \textbf{Initialization}: First of all, $N$ vectors in the size of $|features|$ are initialized with 0.5 as the set of PVs. In fact, 0.5 indicates the probability of selection of each feature at the beginning, which equals the elimination chance. A random population of size $N$ is also generated. Each individual in the population, also known as a candidate solution, is composed of bits - 0s and 1s. A variable - or bit - determines whether the corresponding feature in the original dataset is selected (represented by 1) or removed (represented by 0). As a result, each individual represents a distinct set of features. After calculating the objective values for each individual, the population is considered as the leaders. NDS is applied to the population to determine the first Pareto front.

   \item \textbf {Updating PV}: As presented in Algorithm~\ref{alg-two}, in order to update PVs, 
    we compute the distance matrix between PVs and leaders. As mentioned previously, leaders are $N$ best candidate solutions. To compute the hamming distance between each leader and PV, we need to convert PVs to binary vectors. A PV cell with a value greater than 0.5 indicates the selection of the feature, thus we change it to 1 and a value less than 0.5 indicates that a feature is not selected, thus it becomes zero. After calculating the distance matrix, we assign each leader to the nearest PV.  Each leader is tied to exactly one PV, and each PV is updated using the closest leader. Instead of comparing two individuals, we only consider the leader's variable to update a PV element. If the $j$\textsuperscript{th} variable of the leader assigned to the $i$\textsuperscript{th} PV has a value of 1, then the step size will be added to the corresponding PV element: $PV[i][j] = PV[i][j] + stepSize$, and if that variable's value equals 0 then $PV[i][j] = PV[i][j] - stepSize$. The $stepSize$ is a control parameter that governs the pace at which a PV gets updates.

    To avoid PV elements going below 0 or above 1, we clip these values to add a chance of a mutation to each variable. This is done using the $MinBoundry$ parameter, which controls the interval of PV variables and is the minimum value for each PV element. In the same way, $1-MinBoundry$ is the maximum value. $MinBoundry > 0$ ensures that when generating an individual, each variable can mutate against the elite and we may avoid premature convergence by this means.

   \item \textbf {Generation of New Candidate Solutions}: One new individual is sampled from each PV and added to the population. To this end, based on the probability given for each variable, the status of the corresponding feature is randomly set. Then, each new feature subset should be evaluated based on both objectives. 

   \item \textbf {Selection}: NDS is applied to the population to determine the Pareto front. After calculating the crowding distance of the individuals, the $N$ best individuals are selected as leaders similar to the NSGA-II algorithm. Individuals other than the leaders and the Pareto front can then be eliminated to conserve memory. If the population size exceeds a certain threshold, we select the best solutions and eliminate the others from the population.
   If the termination condition is not met, the next iteration starts by jumping to Step 2.
\end{enumerate}

Individuals that are neither on the Pareto front nor leaders will be eliminated from the population since we do not need them in the following steps. As a result, the number of individuals at the end of each iteration equals $max(N, |Pareto\: front|)$. Before the selection process, when maximum memory is needed, this number is $max(N, |Pareto\: front|) + N$, while NSGA-II requires the memory required for $2 \times Population\:size$ individuals, which is usually much larger than for CNSGA-II since $N$ and $|Pareto\: front|$ are both less than $Population\:size$.

\begin{algorithm2e}
\SetAlgoLined
\SetKwInOut{Input}{input}\SetKwInOut{Output}{output}
 \Input{ $dataset$, $N$, $maxIteration$, $MinBoundry$, $stepSize$, $maxPOPSize$ }
 \tcp{$N = number \: of \: PVs$}
 \Output{ $Pareto front$ }
 \BlankLine
\tcp{Initialization}
$PVs$ = $N$ vectors of size $NumberofFeatures$ with the initial value of $0.5$ for each element\;
$population$ = random population with size $N$\;
evaluate($population$)\;
$leaders = population$\;
$ParetoFront = NDS(population).front[0]$\;

\For{$i\leftarrow 1$ \KwTo $maxIteration$}{

Update PVs\;

\tcp*[h]{Generating new individuals}\; 
\For{$j\leftarrow 1$ \KwTo $N$}
    {
    \For{$k\leftarrow 1$ \KwTo $NumberofFeatures$}
    {
        \eIf{$randomNumber(0,1) < PVs[j][k]$}
        {$NewIndividual[k] = 1$\;}
        {$NewIndividual[k] = 0$\;}
    }
    evaluate($NewIndividual$)\;
    $population = NewIndividual \cup population$\;
    }

$ParetoFront = NDS(population).front[0]$\;

$leaders = N$ best individuals of the population\;
$population = ParetoFront \cup leaders$\;
\If{$len(population)>maxPOPSize$} {
$population = maxPOPSize$ best individuals of the population\;}
}
\BlankLine
\caption{Pseudo-code of binary CNSGA-II}\label{alg-one}
\end{algorithm2e}

\begin{algorithm2e}
\SetAlgoLined
\SetKwInOut{Input}{input}\SetKwInOut{Output}{output}
 \Input{ $PVs$, $dataset$, $N$, $MinBoundry$, $stepSize$ }
 \tcp{$N = number \: of \: PVs$}
 \tcp{$stepSize = 1/virtual \: population\:size$}
 \Output{ $PVs$ }
 \BlankLine

\tcp{Converting PVs to binary vectors}
\For{$j\leftarrow 1$ \KwTo $N$}
    {
    \For{$k\leftarrow 1$ \KwTo $NumberofFeatures$}
        {
        \eIf{$PVs[j][k] > 0.5$}
        {$BinaryPVs[j][k] = 1$}
        {$BinaryPVs[j][k] = 0$} 
        }
    }
\tcp{Calculating Distance Matrix (DM)}
\For{$j\leftarrow 1$ \KwTo $N$}
    {
    \For{$k\leftarrow 1$ \KwTo $N$}
        {
        $DM[j][k] = $ Hamming distance($leaders[k]$, $BinaryPVs[j])$\;
        }
    }
\tcp{Assigning leaders and updating PVs}
\For{$j\leftarrow 1$ \KwTo $N$}
    {
    Assign the nearest leader among unassigned leaders to PV[$j$] and remove it from $leaders$\;
    \For{$k\leftarrow 1$ \KwTo $NumberofFeatures$}
        {
        \eIf{$AssignedLeader[j][k] == 1$}
        {$PVs[j][k] += stepSize$}
        {$PVs[j][k] -= stepSize$}
        }
    }
$PVs = PVs.clip(MinBoundry, 1 - MinBoundry)$\;

\BlankLine
\caption{PV Update Procedure}\label{alg-two}
\end{algorithm2e}

\section{EXPERIMENTAL RESULTS}

\subsection{Datasets}
To evaluate the performance of CNSGA-II for feature selection, we have compared it with NSGA-II on five large-scale datasets in the fields of microarray and image or face recognition \cite{zhao2010advancing}. These datasets are characterized by high feature counts. Due to this, classifier performance deteriorates and irrelevant and redundant features must be removed. Detailed information on the adopted datasets is provided in Table~\ref{tab-datasets}.

\begin{table*}[htbp]
\caption{Datasets Description}
\begin{center}
\begin{tabular}{|c|c|c|c|c|}
\hline
\textbf{Dataset}     & \textbf{\#Features} & \textbf{\#Instances} & \textbf{\#Classes} & \textbf{Domain}          \\ \hline
warpAR10P   & 2400       & 130         & 10        & Image, Face     \\ \hline
warpPIE10P  & 2420       & 210         & 10        & Image, Face     \\ \hline
TOX-171     & 5748       & 171         & 4         & Microarray \\ \hline
pixraw10P   & 10000      & 100         & 10        & Image, Face     \\ \hline
CLL-SUB-111 & 11340      & 111         & 3         & Microarray \\ \hline
\end{tabular}
\label{tab-datasets}
\end{center}
\end{table*}

\subsection{Experimental Settings}
To evaluate the algorithms' performance, the data instances are divided into two subsets: about twenty percent of each dataset's samples are randomly chosen as the test set and the rest of the data is used to train the optimizer. Test set instances are not seen during optimization. Each algorithm is run 10 times on all datasets, and at each run, the train and test sets are distinctly selected. As a result, the experiment can be considered a 10-time hold-out or random subsampling.
Inspired by a similar experiment conducted by Xu et al. \cite{xu2020duplication} we have fixed the number of iterations to 100 for NSGA-II, therefore we have 10,000 function calls to evaluate both algorithms fairly. Details of the hyperparameter settings are provided in Table~\ref{tab-settings}. 

We use hypervolume (HV) as the multi-objective performance evaluation metric to compare optimization algorithms. $k$-Nearest Neighbor ($k$-NN) is employed to classify the data based on the selected features and compute the classification error as the first objective. The $k$-NN implementation with FAISS \cite{fastKNN} is chosen as the classifier to compute the classification error for each individual. The $k$ parameter of $k$-NN for warpAR10P, warpPIE10P, TOX-171, pixraw10P, and CLL-SUB-111 datasets is 5, 5, 5, 5, and 4, respectively.

\begin{table}[htbp]
\caption{Parameter Settings of CNSGA-II and NSGA-II algorithms}
\begin{center}
\begin{tabular}{|ll|}
\hline
\multicolumn{2}{|c|}{\textbf{CNSGA-II}}                                       \\ \hline
\multicolumn{1}{|l|}{Number of PVs (N)}              & 10                     \\ \hline
\multicolumn{1}{|l|}{Step size}                      & 1/500                 \\ \hline
\multicolumn{1}{|l|}{Number of iterations}           & 1000                   \\ \hline
\multicolumn{1}{|l|}{Number of function calls (NFC)} & 10000                  \\ \hline
\multicolumn{1}{|l|}{MinBoundry   }                  & 0.01                   \\ \hline
\multicolumn{1}{|l|}{Maximum size of Pareto front}   & 100                    \\ \hline
\multicolumn{1}{|l|}{Sampling method}                & Binary Random Sampling \\ \hline
\multicolumn{1}{|l|}{Survival method}                & NDS algorithm          \\ \hline
\multicolumn{1}{|l|}{Duplicate Elimination}          & TRUE                   \\ \hline
\multicolumn{1}{|l|}{Number of runs of algorithm}    & 10                     \\ \hline
\multicolumn{2}{|c|}{\textbf{NSGA-II}}                                        \\ \hline
\multicolumn{1}{|l|}{Population size}                & 100                    \\ \hline
\multicolumn{1}{|l|}{Number of iterations}           & 100                    \\ \hline
\multicolumn{1}{|l|}{Number of function calls (NFC)} & 10000                  \\ \hline
\multicolumn{1}{|l|}{Sampling method}                & Binary Random Sampling \\ \hline
\multicolumn{1}{|l|}{Selection method}               & Binary Tournament      \\ \hline
\multicolumn{1}{|l|}{Mutation method}                & Bit-flip Mutation      \\ \hline
\multicolumn{1}{|l|}{Mutation probability}           & $1/dimension$          \\ \hline
\multicolumn{1}{|l|}{Crossover method}               & SPX                    \\ \hline
\multicolumn{1}{|l|}{Crossover probability}          & 1.0                    \\ \hline
\multicolumn{1}{|l|}{Survival method}                & NDS algorithm          \\ \hline
\multicolumn{1}{|l|}{Duplicate Elimination}          & TRUE                   \\ \hline
\multicolumn{1}{|l|}{Number of runs of algorithm}    & 10                     \\ \hline
\end{tabular}
\label{tab-settings}
\end{center}
\end{table}

\subsection{Numerical Results and Analysis}
\begin{figure*}
\centering
\begin{tabular}{cc}
\includegraphics[width=0.32\linewidth]{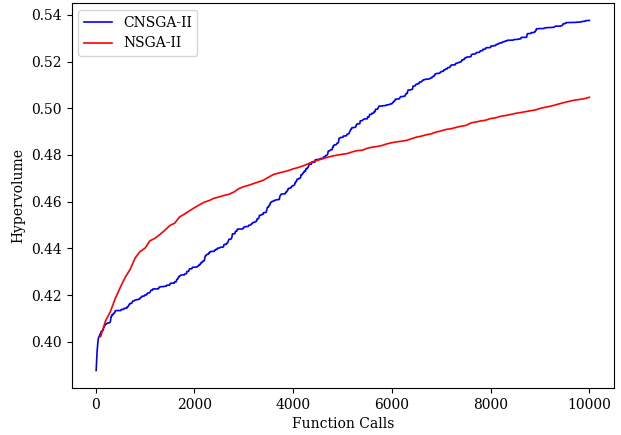} &\includegraphics[width=0.30\linewidth]{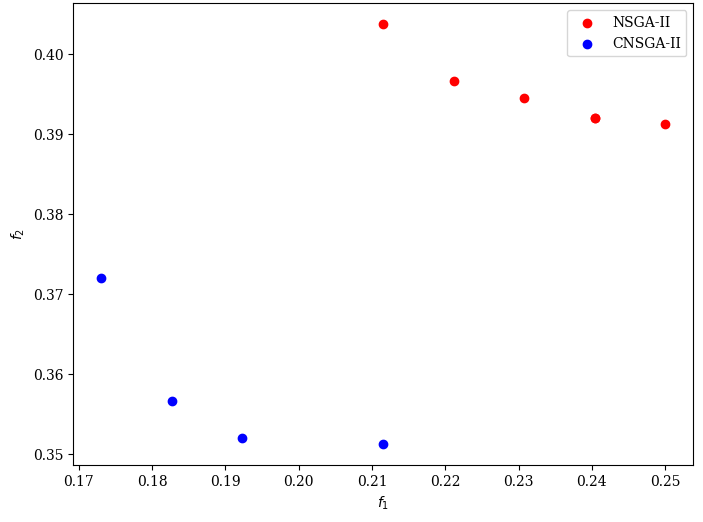}\\
(a) warpAR10P & (b) warpAR10P\\
\includegraphics[width=0.32\linewidth]{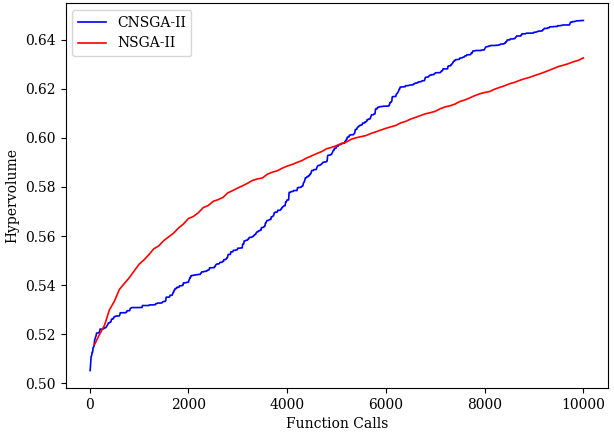} &\includegraphics[width=0.31\linewidth]{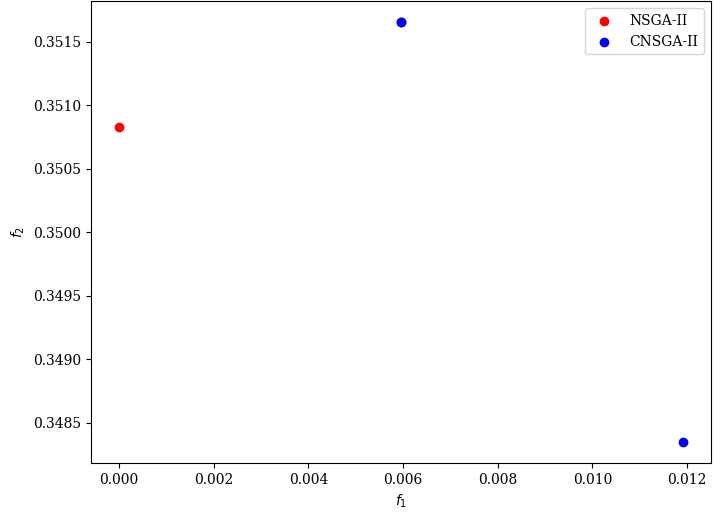}\\
(c) warpPIE10P & (d) warpPIE10P\\
\includegraphics[width=0.32\linewidth]{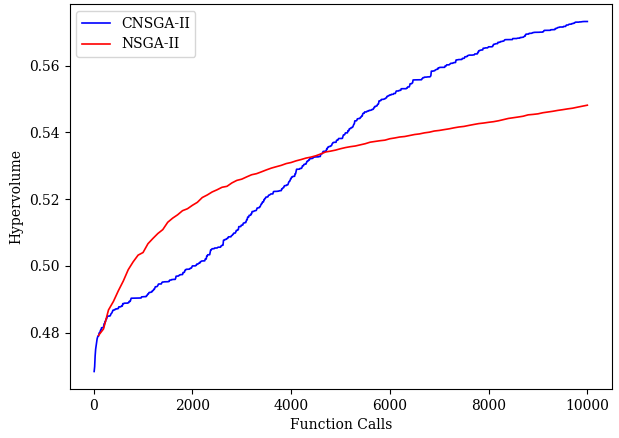} &\includegraphics[width=0.31\linewidth]{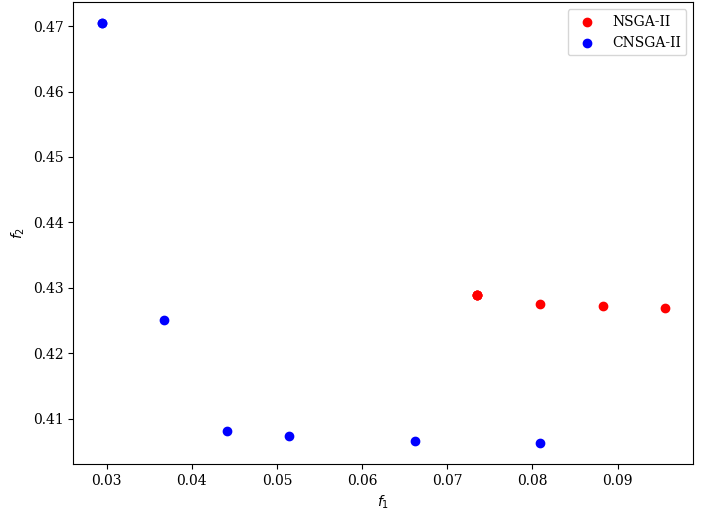}\\
(e) TOX-171 & (f) TOX-171\\
\includegraphics[width=0.32\linewidth]{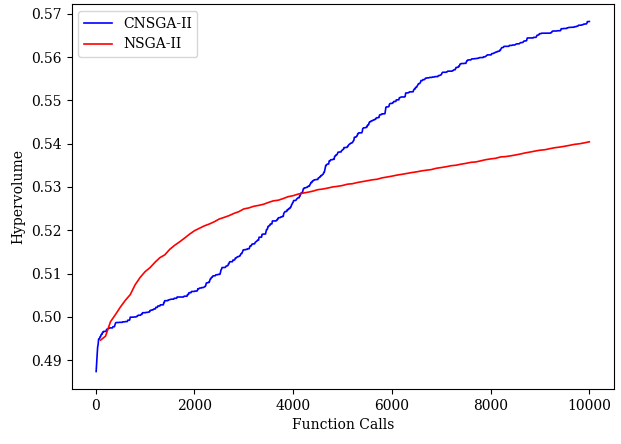} &\includegraphics[width=0.31\linewidth]{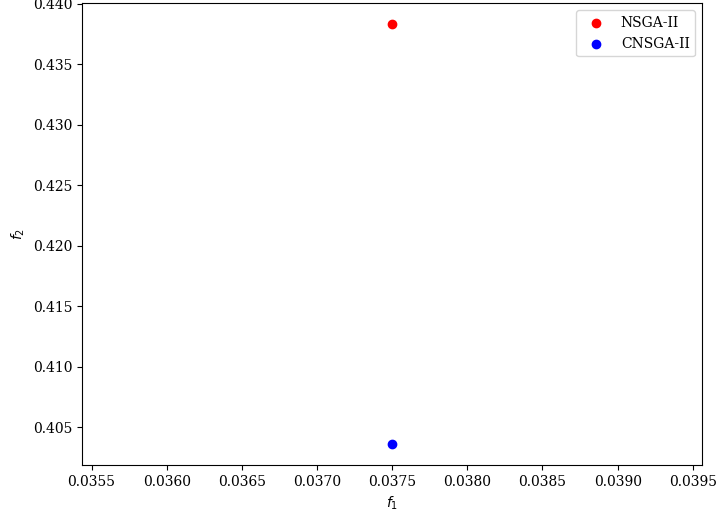}\\
(g) pixraw10P & (h) pixraw10P\\
\includegraphics[width=0.32\linewidth]{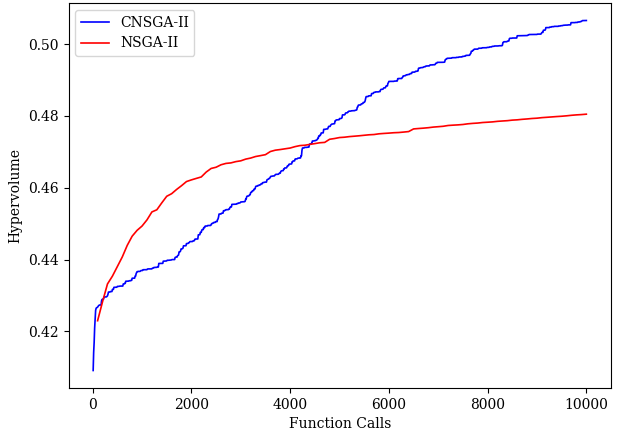} &\includegraphics[width=0.31\linewidth]{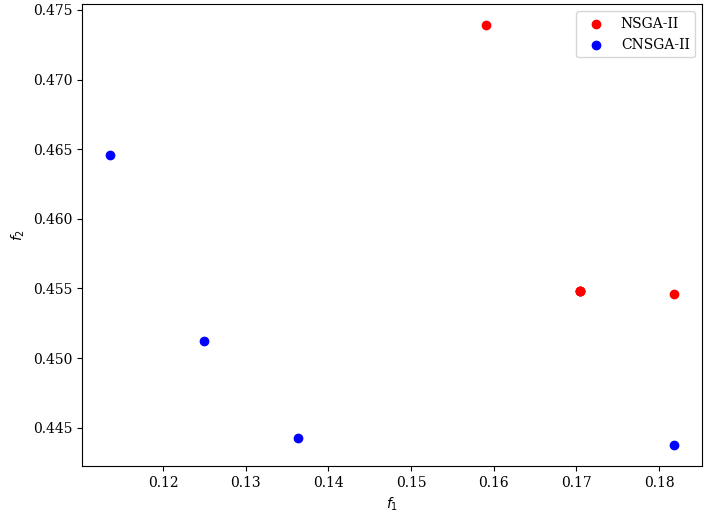}\\
(i) CLL-SUB-111 & (j) CLL-SUB-111\\

\end{tabular}
\caption{HV plots during optimization with $StepSize=1/500$ on the train set (left) and median final train Pareto set (right) on datasets. $f_1$ and $f_2$ are  the classification error and the ratio of selected features, respectively.}
\label{img-500-par}
\end{figure*}

Fig. \ref{img-500-par} represents the average trend of the HV during optimization on the left side and the final Pareto fronts on the train sets on the right side. For all the datasets in the mentioned settings, NSGA-II reaches a lower train HV value than the compact method in these large-scale search spaces. While NSGA-II's HV rises faster at the beginning iterations, CNSGA-II eventually surpasses that. CNSGA-II's superior performance is achieved using less memory. In addition, in 4 datasets out of 5, the train Pareto front solutions resulting from CNSGA-II dominate all the solutions resulting from NSGA-II, and the distribution of the solutions on CNSGA-II's Pareto front is at least as good as NSGA-II's.

Table~\ref{tab-results-HV} presents the average numerical results for the final Pareto fronts obtained from CNSGA-II and NSGA-II algorithms. We notice that, for all datasets, CNSGA-II's train and test HVs are superior to NSGA-II. 
In both algorithms, the train HV increases slower for a large dataset such as CLL-SUB-111, and it requires more function calls. We have also reported the average HV values on the initial test sets to demonstrate the improvement of HV values during optimization. On average, NSGA-II has improved the HV of the test sets by about 0.05, while CNSGA-II with an improvement of about 0.07 has proved its supremacy. For the largest dataset, CLL-SUB-111, the improvement of HV on CNSGA-II's test Pareto front is more than 2 times the improvement of HV on NSGA-II's test Pareto front.

Table~\ref{tab-results-Obj} represents the minimum classification error of the train set and the average ratio of selected features on the final Pareto fronts for CNSGA-II and NSGA-II. Both algorithms have reduced the number of features, and accordingly the data size, by more than 50\%. However, we see that CNSGA-II is more successful at reducing the ratio of selected features than NSGA-II (second objective) for all datasets. In addition, CNSGA-II is the winning algorithm in minimizing the classification error of the train set for 3 out of 5 datasets by at least 0.04. For the rest, both algorithms are almost comparable.

\begin{figure*}
\centering
\begin{tabular}{cc}
\includegraphics[width=0.42\linewidth]{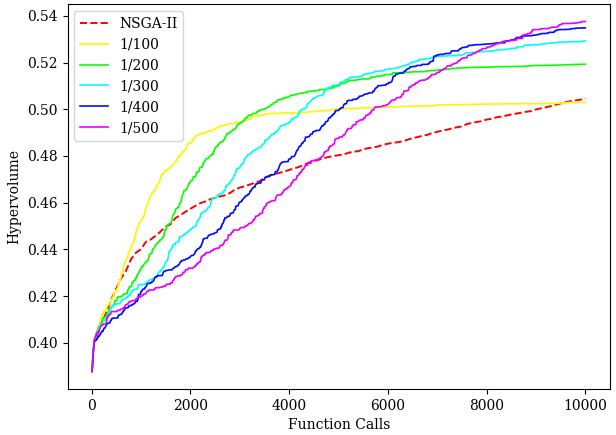} &\includegraphics[width=0.42\linewidth]{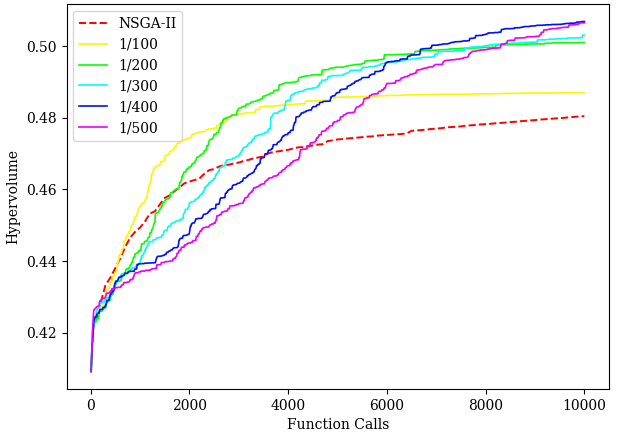}\\
(a) warpAR10P & (b) CLL-SUB-111\\

\end{tabular}
\caption{Comparing the performance of CNSGA-II with various step sizes}
\label{img-comb}
\end{figure*}

One key hyperparameter in CNSGA-II is the step size which governs the pace at which the PVs update. If its value is decreased (near 0), it usually leads to a smoother slope in the train HV trend, but a later convergence and a better final HV, as well as better Pareto front solutions when more budget is available. On the other hand, when we set the step size to a larger value (near 1), HV rises faster but converges prematurely at a smaller HV value. In other words, with a large step size, the optimizer falls into local optima. Fig.~\ref{img-comb} compares the algorithm's performance with various step sizes. We believe that this hyperparameter makes the method more flexible for different applications. When we need to solve an expensive multi-objective problem with a very limited number of evaluation calls, CNSGA-II with a large step size results in a better Pareto front than NSGA-II. With more budget available, the step size can be set smaller. 

Besides, the number of PVs ($N$) plays a critical role in memory usage and the quality of the solutions on the Pareto front. By reducing $N$ less memory is used but we may miss some potential areas in the search space. Tuning this hyperparameter depends on the application and available memory size.

\begin{table*}[htbp]
\caption{Results to compare average final train and test HV of CNSGA-II and NSGA-II. Initial HV is computed over the initial test sets.}
\begin{center}
\begin{tabular}{|c|c|cc|cc|}
\hline
                 &                     & \multicolumn{2}{c|}{\textbf{CNSGA-II}}                                & \multicolumn{2}{c|}{\textbf{NSGA-II}}                                 \\ \hline
\textbf{Dataset} & \textbf{Initial HV} & \multicolumn{1}{c|}{\textbf{Final Train HV}} & \textbf{Final Test HV} & \multicolumn{1}{c|}{\textbf{Final Train HV}} & \textbf{Final Test HV} \\ \hline
warpAR10P        & 0.26±0.03           & \multicolumn{1}{c|}{\textbf{0.54±0.03}}      & \textbf{0.33±0.04}     & \multicolumn{1}{c|}{0.50±0.03}               & 0.32±0.04              \\ \hline
warpPIE10P       & 0.34±0.04           & \multicolumn{1}{c|}{\textbf{0.65±0.01}}      & \textbf{0.43±0.04}     & \multicolumn{1}{c|}{0.63±0.01}               & 0.41±0.03              \\ \hline
TOX-171          & 0.37±0.04           & \multicolumn{1}{c|}{\textbf{0.57±0.01}}      & \textbf{0.44±0.05}     & \multicolumn{1}{c|}{0.55±0.01}               & 0.41±0.05              \\ \hline
pixraw10P        & 0.35±0.06           & \multicolumn{1}{c|}{\textbf{0.57±0.01}}      & \textbf{0.40±0.07}     & \multicolumn{1}{c|}{0.54±0.01}               & 0.38±0.07              \\ \hline
CLL-SUB-111      & 0.39±0.03           & \multicolumn{1}{c|}{\textbf{0.51±0.01}}      & \textbf{0.44±0.05}     & \multicolumn{1}{c|}{0.48±0.01}               & 0.41±0.04              \\ \hline
\textbf{Average} & 0.34±0.04           & \multicolumn{1}{c|}{\textbf{0.57±0.01}}      & \textbf{0.41±0.05}     & \multicolumn{1}{c|}{0.54±0.01}               & 0.39±0.05              \\ \hline
\end{tabular}
\label{tab-results-HV}
\end{center}
\end{table*}

\begin{table*}[htbp]
\caption{Results for comparing average final objective values in the Pareto fronts of CNSGA-II and NSGA-II}
\begin{center}\
\begin{tabular}{|c|cc|cc|}
\hline
\textbf{}        & \multicolumn{2}{c|}{\textbf{CNSGA-II}}                                                                                                                                              & \multicolumn{2}{c|}{\textbf{NSGA-II}}                                                                                                                                               \\ \hline
\textbf{Dataset} & \multicolumn{1}{c|}{\textbf{\begin{tabular}[c]{@{}c@{}}Minimum\\ Classification Error\end{tabular}}} & \textbf{\begin{tabular}[c]{@{}c@{}}Average\\ Ratio of Features\end{tabular}} & \multicolumn{1}{c|}{\textbf{\begin{tabular}[c]{@{}c@{}}Minimum\\ Classification Error\end{tabular}}} & \textbf{\begin{tabular}[c]{@{}c@{}}Average\\ Ratio of Features\end{tabular}} \\ \hline
warpAR10P        & \multicolumn{1}{c|}{\textbf{0.16±0.04}}                                                              & \textbf{0.36±0.01}                                                           & \multicolumn{1}{c|}{0.21±0.01}                                                                       & 0.38±0.01                                                                    \\ \hline
warpPIE10P       & \multicolumn{1}{c|}{0.01±0.01}                                                                       & \textbf{0.35±0.01}                                                           & \multicolumn{1}{c|}{0.01±0.01}                                                                       & 0.36±0.01                                                                    \\ \hline
TOX-171          & \multicolumn{1}{c|}{\textbf{0.03±0.01}}                                                              & \textbf{0.42±0.01}                                                           & \multicolumn{1}{c|}{0.07±0.01}                                                                       & 0.43±0.01                                                                    \\ \hline
pixraw10P        & \multicolumn{1}{c|}{0.03±0.01}                                                                       & \textbf{0.41±0.01}                                                           & \multicolumn{1}{c|}{0.03±0.01}                                                                       & 0.44±0.01                                                                    \\ \hline
CLL-SUB-111      & \multicolumn{1}{c|}{\textbf{0.10±0.02}}                                                              & \textbf{0.44±0.01}                                                           & \multicolumn{1}{c|}{0.16±0.01}                                                                       & 0.45±0.01                                                                    \\ \hline
\textbf{Average} & \multicolumn{1}{c|}{\textbf{0.07±0.02}}                                                              & \textbf{0.40±0.01}                                                           & \multicolumn{1}{c|}{0.10±0.01}                                                                       & 0.41±0.01                                                                    \\ \hline
\end{tabular}
\label{tab-results-Obj}
\end{center}
\end{table*}

\section{CONCLUSION REMARKS}

Feature selection is a critical pre-processing task in big data analysis since computational requirements grow with data size. To minimize the data dimension and classification error of real-world large-scale datasets, we propose CNSGA-II, which uses less memory than the famous NSGA-II and performs better in terms of the final hyper-volume and minimizing the objectives. To the best of our knowledge, it is the first compact multi-objective method applied to feature selection. In this method, new individuals are generated using sampling from probability vectors. These vectors are updated iteratively based on the best candidate solutions in the population. According to the experimental results, CNSGA-II is more effective at improving classification accuracy and reducing features compared to NSGA-II.

As a suggestion for future work, CNSGA-II can be tested on huge-scale datasets, many-objective problems, and multi-objective problems other than feature selection. More exploration can improve the PV-update function. It is also possible to design and implement a real-value scheme of this method.

\bibliography{ref}

\begin{thebibliography}{10}
\providecommand{\url}[1]{#1}
\csname url@samestyle\endcsname
\providecommand{\newblock}{\relax}
\providecommand{\bibinfo}[2]{#2}
\providecommand{\BIBentrySTDinterwordspacing}{\spaceskip=0pt\relax}
\providecommand{\BIBentryALTinterwordstretchfactor}{4}
\providecommand{\BIBentryALTinterwordspacing}{\spaceskip=\fontdimen2\font plus
\BIBentryALTinterwordstretchfactor\fontdimen3\font minus \fontdimen4\font\relax}
\providecommand{\BIBforeignlanguage}[2]{{%
\expandafter\ifx\csname l@#1\endcsname\relax
\typeout{** WARNING: IEEEtran.bst: No hyphenation pattern has been}%
\typeout{** loaded for the language `#1'. Using the pattern for}%
\typeout{** the default language instead.}%
\else
\language=\csname l@#1\endcsname
\fi
#2}}
\providecommand{\BIBdecl}{\relax}
\BIBdecl

\bibitem{chandrashekar2014survey}
G.~Chandrashekar and F.~Sahin, ``A survey on feature selection methods,'' \emph{Computers \& Electrical Engineering}, vol.~40, no.~1, pp. 16--28, 2014.

\bibitem{ghalwash2016structured}
M.~F. Ghalwash, X.~H. Cao, I.~Stojkovic, and Z.~Obradovic, ``Structured feature selection using coordinate descent optimization,'' \emph{BMC bioinformatics}, vol.~17, no.~1, pp. 1--14, 2016.

\bibitem{bidgoli2020evolutionary}
A.~A. Bidgoli, H.~Ebrahimpour-Komleh, and S.~Rahnamayan, ``An evolutionary decomposition-based multi-objective feature selection for multi-label classification,'' \emph{PeerJ Computer Science}, vol.~6, p. e261, 2020.

\bibitem{hamdani2007multi}
T.~M. Hamdani, J.-M. Won, A.~M. Alimi, and F.~Karray, ``Multi-objective feature selection with nsga ii,'' in \emph{Adaptive and Natural Computing Algorithms: 8th International Conference, ICANNGA 2007, Warsaw, Poland, April 11-14, 2007, Proceedings, Part I 8}.\hskip 1em plus 0.5em minus 0.4em\relax Springer, 2007, pp. 240--247.

\bibitem{deb2002fast}
K.~Deb, A.~Pratap, S.~Agarwal, and T.~Meyarivan, ``A fast and elitist multiobjective genetic algorithm: Nsga-ii,'' \emph{IEEE transactions on evolutionary computation}, vol.~6, no.~2, pp. 182--197, 2002.

\bibitem{jiao2022solving}
R.~Jiao, B.~Xue, and M.~Zhang, ``Solving multi-objective feature selection problems in classification via problem reformulation and duplication handling,'' \emph{IEEE Transactions on Evolutionary Computation}, 2022.

\bibitem{cheng2022variable}
F.~Cheng, J.~J. Cui, Q.~J. Wang, and L.~Zhang, ``A variable granularity search based multi-objective feature selection algorithm for high-dimensional data classification,'' \emph{IEEE Transactions on Evolutionary Computation}, 2022.

\bibitem{harik1999compact}
G.~R. Harik, F.~G. Lobo, and D.~E. Goldberg, ``The compact genetic algorithm,'' \emph{IEEE transactions on evolutionary computation}, vol.~3, no.~4, pp. 287--297, 1999.

\bibitem{aporntewan2001hardware}
C.~Aporntewan and P.~Chongstitvatana, ``A hardware implementation of the compact genetic algorithm,'' in \emph{Proceedings of the 2001 congress on evolutionary computation (ieee cat. no. 01th8546)}, vol.~1.\hskip 1em plus 0.5em minus 0.4em\relax IEEE, 2001, pp. 624--629.

\bibitem{gallagher2004family}
J.~C. Gallagher, S.~Vigraham, and G.~Kramer, ``A family of compact genetic algorithms for intrinsic evolvable hardware,'' \emph{IEEE Transactions on evolutionary computation}, vol.~8, no.~2, pp. 111--126, 2004.

\bibitem{jewajinda2008cellular}
Y.~Jewajinda and P.~Chongstitvatana, ``Cellular compact genetic algorithm for evolvable hardware,'' in \emph{2008 5th International Conference on Electrical Engineering/Electronics, Computer, Telecommunications and Information Technology}, vol.~1.\hskip 1em plus 0.5em minus 0.4em\relax IEEE, 2008, pp. 1--4.

\bibitem{mininno2008real}
E.~Mininno, F.~Cupertino, and D.~Naso, ``Real-valued compact genetic algorithms for embedded microcontroller optimization,'' \emph{IEEE Transactions on Evolutionary Computation}, vol.~12, no.~2, pp. 203--219, 2008.

\bibitem{iacca2019compact}
G.~Iacca and F.~Caraffini, ``Compact optimization algorithms with re-sampled inheritance,'' in \emph{Applications of Evolutionary Computation: 22nd International Conference, EvoApplications 2019, Held as Part of EvoStar 2019, Leipzig, Germany, April 24--26, 2019, Proceedings 22}.\hskip 1em plus 0.5em minus 0.4em\relax Springer, 2019, pp. 523--534.

\bibitem{iacca2020re}
------, ``Re-sampled inheritance compact optimization,'' \emph{Knowledge-Based Systems}, vol. 208, p. 106416, 2020.

\bibitem{velazquez2014multi}
J.~M.~O. Velazquez, C.~A.~C. Coello, and A.~Arias-Montano, ``Multi-objective compact differential evolution,'' in \emph{2014 IEEE symposium on differential evolution (SDE)}.\hskip 1em plus 0.5em minus 0.4em\relax IEEE, 2014, pp. 1--8.

\bibitem{mininno2010compact}
E.~Mininno, F.~Neri, F.~Cupertino, and D.~Naso, ``Compact differential evolution,'' \emph{IEEE Transactions on Evolutionary Computation}, vol.~15, no.~1, pp. 32--54, 2010.

\bibitem{ahn2003elitism}
C.~W. Ahn and R.~S. Ramakrishna, ``Elitism-based compact genetic algorithms,'' \emph{IEEE Transactions on Evolutionary Computation}, vol.~7, no.~4, pp. 367--385, 2003.

\bibitem{harik1999linkage}
G.~Harik \emph{et~al.}, ``Linkage learning via probabilistic modeling in the ecga,'' \emph{IlliGAL report}, vol. 99010, 1999.

\bibitem{harik2006linkage}
G.~R. Harik, F.~G. Lobo, and K.~Sastry, ``Linkage learning via probabilistic modeling in the extended compact genetic algorithm (ecga),'' \emph{Scalable optimization via probabilistic modeling}, pp. 39--61, 2006.

\bibitem{sastry2000extended}
K.~Sastry and D.~E. Goldberg, ``On extended compact genetic algorithm,'' in \emph{Late-Breaking Paper at the Genetic and Evolutionary Computation Conference}, 2000, pp. 352--359.

\bibitem{neri2013compact}
F.~Neri, E.~Mininno, and G.~Iacca, ``Compact particle swarm optimization,'' \emph{Information Sciences}, vol. 239, pp. 96--121, 2013.

\bibitem{zheng2021compact}
W.-M. Zheng, N.~Liu, Q.-W. Chai, and S.-C. Chu, ``A compact adaptive particle swarm optimization algorithm in the application of the mobile sensor localization,'' \emph{Wireless Communications and Mobile Computing}, vol. 2021, pp. 1--15, 2021.

\bibitem{liu2022novel}
N.~Liu, Q.-W. Chai, S.~Liu, W.-M. Zheng \emph{et~al.}, ``A novel compact particle swarm optimization for optimizing coverage of 3d in wireless sensor network,'' \emph{Wireless Communications and Mobile Computing}, vol. 2022, 2022.

\bibitem{asilian2022machine}
A.~Asilian~Bidgoli, S.~Rahnamayan, B.~Erdem, Z.~Erdem, A.~Ibrahim, K.~Deb, and A.~Grami, ``Machine learning-based framework to cover optimal pareto-front in many-objective optimization,'' \emph{Complex \& Intelligent Systems}, vol.~8, no.~6, pp. 5287--5308, 2022.

\bibitem{bidgoli2021reference}
A.~A. Bidgoli, H.~Ebrahimpour-Komleh, and S.~Rahnamayan, ``Reference-point-based multi-objective optimization algorithm with opposition-based voting scheme for multi-label feature selection,'' \emph{Information Sciences}, vol. 547, pp. 1--17, 2021.

\bibitem{zhao2010advancing}
Z.~Zhao, F.~Morstatter, S.~Sharma, S.~Alelyani, A.~Anand, and H.~Liu, ``Advancing feature selection research,'' \emph{ASU feature selection repository}, pp. 1--28, 2010.

\bibitem{xu2020duplication}
H.~Xu, B.~Xue, and M.~Zhang, ``A duplication analysis-based evolutionary algorithm for biobjective feature selection,'' \emph{IEEE Transactions on Evolutionary Computation}, vol.~25, no.~2, pp. 205--218, 2020.

\bibitem{fastKNN}
\BIBentryALTinterwordspacing
J.~Adamczyk. (2020) Make knn 300 times faster than scikit-learn’s in 20 lines! [Online]. Available: \url{https://towardsdatascience.com/make-knn-300-times-faster-than-scikit-learns-in-20-lines-5e29d74e76bb}
\BIBentrySTDinterwordspacing

\end{thebibliography}
\bibliographystyle{IEEEtran}

\end{document}